\documentclass[paper=letter, fontsize=20pt]{article}
\usepackage[square]{natbib}
\usepackage[english]{babel} 
\usepackage{amsmath,amsfonts,amsthm} 
\usepackage[utf8]{inputenc}
\usepackage{float}
\usepackage{lipsum} 
\usepackage{multirow}
\usepackage{array}
\usepackage{blindtext}
\usepackage{graphicx,subfigure}
\usepackage{caption}
\usepackage{appendix}
\makeatletter
\def\@seccntformat#1{\@ifundefined{#1@cntformat}%
   {\csname the#1\endcsname\quad}  
   {\csname #1@cntformat\endcsname}
}
\let\oldappendix\appendix 
\renewcommand\appendix{%
    \oldappendix
    \newcommand{\section@cntformat}{\appendixname~\thesection\quad}
}
\makeatother

\usepackage[sc]{mathpazo} 
\usepackage[T1]{fontenc} 
\usepackage{microtype} 
\usepackage[hmarginratio=1:1,top=32mm,columnsep=20pt]{geometry} 
\usepackage{multicol} 
\usepackage{booktabs} 
\usepackage{float} 
\usepackage{hyperref} 
\usepackage{lettrine} 
\usepackage{paralist} 
\usepackage{abstract} 
\usepackage{titlesec} 
\usepackage{enumitem}
\usepackage{algorithm}
\usepackage{algorithmic}
\renewcommand{\algorithmicrequire}{ \textbf{Input:}} 
\renewcommand{\algorithmicensure}{ \textbf{Output:}}

\title{\vspace{-7mm}\fontsize{17pt}{10pt}\selectfont\textbf{Multi-Label Learning with Deep Forest}} 
\author{
Liang Yang, Xi-Zhu Wu, Yuan Jiang and Zhi-Hua Zhou\\ 
\small
National Key Laboratory for Novel Software Technology,\\
\small
Nanjing University, Nanjing 210023, China\\
\small
\{yangl, wuxz, jiangy, zhouzh\}@lamda.nju.edu.cn\\
}
\date{}

\begin{document}
\maketitle

\hrule
\begin{abstract} 
In multi-label learning, each instance is associated with multiple labels and the crucial task is how to leverage label correlations in building models. 
Deep neural network methods usually jointly embed the feature and label information into a latent space to exploit label correlations. 
However, the success of these methods highly depends on the precise choice of model depth.
Deep forest is a recent deep learning framework based on tree model ensembles, which does not rely on backpropagation.
We consider the advantages of deep forest models are very appropriate for solving multi-label problems.
Therefore we design the Multi-Label Deep Forest (MLDF) method with two mechanisms: measure-aware feature reuse and measure-aware layer growth. 
The measure-aware feature reuse mechanism reuses the good representation in the previous layer guided by confidence.
The measure-aware layer growth mechanism ensures MLDF gradually increase the model complexity by performance measure.
MLDF handles two challenging problems at the same time: one is restricting the model complexity to ease the overfitting issue; another is optimizing the performance measure on user's demand since there are many different measures in the multi-label evaluation. 
Experiments show that our proposal not only beats the compared methods over six measures on benchmark datasets but also enjoys label correlation discovery and other desired properties in multi-label learning.

\end{abstract} 
\hrule

\section{Introduction}
In multi-label learning, each example is associated with multiple labels simultaneously and the task of multi-label learning is to predict a set of relevant labels for the unseen instance. Multi-label learning has been widely applied in diverse problems like text categorization \citep{zhou2015ontology}, scene classification \citep{zhao2015deep}, functional genomics \citep{DBLP:journals/tkde/ZhangZ06},  video categorization \citep{ray2018scenes}, chemicals classification \citep{cheng2017iatc}, etc.
It is drawn increased research attention that multi-label learning tasks are omnipresent in real-world problems \citep{mining}.

By transforming the multi-label learning problem to independent binary classification problems for each label, Binary Relevance \citep{BR} is a straightforward method which has been widely used in practice. 
Though it aims to make full use of high-performance traditional single-label classifiers, it will lead to high computational cost when label space is large.
Besides, such a method neglects the fact that information on one label may be helpful for learning other related labels, which would limit the prediction performance.
Investigating correlations among labels has been demonstrated to be the key to improve the performance of multi-label learning.
As a result, more and more multi-label learning methods aim to explore and exploit the label correlations are proposed \citep{mining}.
There emerges considerable attention in exploring and exploiting label correlations in multi-label learning methods \citep{mining, wang2016cnn, ZhangZZ18}. 

Different from traditional multi-label methods, deep neural network models usually try to learn a new feature space and employ a multi-label classifier on the top.
Among the first to utilize network architectures, BP-MLL \citep{DBLP:journals/tkde/ZhangZ06} not only treats each output node as a binary classification task but also exploits label correlations relied on the architecture itself.
Later, a comparably simple Neural Network approach builds upon BP-MLL was proposed by replacing the pairwise ranking loss with entropy loss and using deep neural networks techniques \citep{NN_pro} and it achieves a good result in the large-scale text classification.
However, deep neural models usually require a huge amount of training data, and thus they are generally not suitable for small-scale datasets.

By realizing that the essence of deep learning lies in layer-by-layer processing, in-model feature transformation, and sufficient model complexity, Zhou and Feng proposed deep forest \citep{deepforest}. 
Deep forest is an ensemble deep model built on decision trees and does not use backpropagation in the training process.
A deep forest ensembled with a cascade structure is able to do representation learning similarly like deep neural models. 
Deep forest is much easier to train since it has fewer hyper-parameters.
It has achieved excellent performance on a broad range of tasks, such as large-scale financial fraud detection \citep{zhang2018distributed}, image, and text reconstruction \citep{feng2017autoencoder}.
Though deep forest has been found useful in traditional classification tasks \citep{deepforest}, the potential of applying it into multi-label learning has not been noticed before our work.

The success of deep forest mainly comes from the layer-by-layer feature transformation in an ensemble way \citep{zhou2012ensemble}. 
While on the other hand, the key point in multi-label learning is how to use label correlations. 
Inspired by these two facts, we propose the Multi-Label Deep Forest (MLDF) method. 
Briefly speaking, MLDF uses different multi-label tree methods as the building blocks in deep forest, and label correlations can be exploited via layer-by-layer representation learning. 
Because evaluation in multi-label learning is more complicated than traditional classification tasks, a number of performance measures have been proposed \citep{schapire2000boostexter}. 
It is noticed that ifferent user has different demands and an algorithm usually performs differently on different measures \citep{limo}.
In order to achieve better performance on the specific measure, we propose two mechanisms: measure-aware feature reuse and measure-aware layer growth.
The measure-aware feature reuse mechanism, inspired by confidence screening \citep{pang1501improving}, reuses good presentation in the previous layer.
The measure-aware layer growth mechanism aims to control the model complexity by various performance measures.
The main contributions of this paper are summarized as follows:

\begin{compactitem} 
    \item 
    We first introduce deep forest to multi-label learning. 
	Because of the proposed cascade structure and two measure-aware mechanisms, our MLDF method can handle two challenging problems in multi-label learning at the same time: optimizing different performance measures on user demand, and reducing overfitting, when utilizing label correlations by a large number of layers, which often observed in deep neural multi-label models.
    \item
    Our extensive experiments show that MLDF achieves the best performance on 9 benchmark datasets over 6 multi-label measures. 
	Besides, the two mechanisms are confirmed necessary in MLDF.
	Furthermore, investigative experiments demonstrate our proposal enjoys high flexibility in applying various base tree models and resistance to overfitting.
\end{compactitem} 

The remainder of the paper is organized as follows. Section \ref{preliminaries} introduces some preliminaries. Section \ref{method} formally describes our MLDF method and two measure-aware mechanisms. Section \ref{experiments} reports the experimental results on benchmarks and investigative experiments. Finally, we conclude the paper in Section \ref{conclusion}.

\section{PRELIMINARIES}
\label{preliminaries}
In this section, we introduce various performance measures, followed by the tree-based multi-label methods, which are the base of Multi-Label Deep Forest.

\subsection{Multi-label performance measures}
The multi-label classification task is to derive a function $H$ from the training set $\{(\boldsymbol{x}_i,\boldsymbol{y}_i)|1 \le i \le m, \boldsymbol{x}_i \in \mathbb{R}^d, \boldsymbol{y}_i \in \{0,1\}^l\}$.
Suppose the multi-label learning model first produces a real-valued function $F:\mathcal{X} \rightarrow [0,1]^l$, which can be viewed as the confidence of relevance of the labels. 
The multi-label classifier $H:\mathcal{X} \rightarrow  \{0,1\}^l$ can be induced from $F$ by thresholding.

There are lots of performance measures for multi-label learning.
Six widely-used evaluation measures are employed in this paper \citep{limo}.
Table \ref{measures} shows the formulation of these measures, being $Y$ the true labels, $Y_{i\cdot}$ the $i$-th row of the label matrix, `$+$'(`$-$') the relevant (irrelevant) note.
Hamming loss and macro-AUC are label-based measures, while one-error, coverage, ranking loss and average precision are instance-based measures \citep{zhang2014review}.
The $f_{ij}$ means the confidence score of $i$-th instance on $j$-th label, $h_{ij}$ means the predict result of $i$-th instance on $j$-th label, and $\mathrm{rank}_{ F } \left( \boldsymbol{ x } _ { i } , j \right)$ means the instance $\boldsymbol{ x } _ { i }$'s rank on $j$-th label.
For example, $\boldsymbol{f}(\boldsymbol{x_i})=[0.2,0.8,0.4]$, then we have $\boldsymbol{h}(\boldsymbol{x_i})=[0,1,0]$ with threshold 0.5.
Furthermore, we have $Y _ { i \cdot  } ^ { - }=\{1,3\}$ and $Y _ { i \cdot } ^ { + }=\{2\}$.
Since $f_{i2}=0.8$, then we have $\mathrm{rank}_{ F } ( \boldsymbol{ x } _ { i } , 2)=1$.

\begin{table}
	\caption{Definitions of six multi-label performance measures, `$\downarrow$' means the lower the better, `$\uparrow$' means the higher the better.}
	\label{measures}
	\renewcommand\arraystretch{0.7}	
	\begin{center}
		\begin{tabular}{p{3.0cm}<{\raggedright}l}
			\toprule
			Measure & Formulation \\
			\midrule
			\multirow{3}{*}{\small   hamming loss $\downarrow$} 
			& \multirow{3}{*}{$\frac { 1 } { m l } \sum \limits_{ i = 1 } ^ { m } \sum \limits_{ j = 1 } ^ { l } \mathbb{I}[h _ { i j } \neq y _ { i j } ]$}
			\\
			\multirow{3}{*}{} & \multirow{3}{*}{} \\
			\multirow{3}{*}{} & \multirow{3}{*}{} \\
			\multirow{3}{*}{\small one-error $\downarrow$} 
			& \multirow{3}{*}{\small $\frac { 1 } { m } \sum \limits_ { i = 1 } ^ { m } \mathbb{I}\left[ \arg \max F \left( \boldsymbol{ x } _ { i } \right) \notin Y _ { i \cdot} ^ { + } \right]$}
			\\
			\multirow{3}{*}{} & \multirow{3}{*}{} \\
			\multirow{3}{*}{} & \multirow{3}{*}{} \\
			\multirow{3}{*}{\small   coverage $\downarrow$} 
			& \multirow{3}{*}{\small $\frac{1}{ml}\sum \limits_ { i = 1 } ^ { m } \mathbb{I}\left[ \max \limits_{ j \in Y _ { i \cdot} ^ { + } }  \mathrm{rank}_{ F } \left( \boldsymbol{ x } _ { i } , j \right) - 1 \right]$}
			\\
			\multirow{3}{*}{} & \multirow{4}{*}{} \\
			\multirow{3}{*}{} & \multirow{4}{*}{} \\
			\multirow{3}{*}{\small   ranking loss $\downarrow$}  
			& \multirow{3}{*}{\small $\frac { 1 } { m } \sum \limits_ { i = 1 } ^ { m } \frac { \left| \mathcal { S } _ { \mathrm { rank } } ^ { i } \right| } { \left| Y _ { i\cdot } ^ { + } \right| \left| Y _ { i\cdot } ^ { - } \right| }$}
			\\
			\multirow{3}{*}{} & \multirow{3}{*}{} \\
			\multirow{3}{*}{} & \multirow{3}{*}{} \\
			\multirow{3}{*}{\small   average precision $\uparrow$}
			& \multirow{3}{*}{\small $\frac { 1 } { m } \sum \limits_ { i = 1 } ^ { m } \frac { 1 } { \left| Y _ { i \cdot} ^ { + } \right| } \sum \limits_{ j \in Y _ { i \cdot } ^ { + } } \frac { \left| \mathcal { S } _ {\mathrm{precision}} ^ { i j } \right| } { \mathrm{rank} _ { F } \left( \boldsymbol{ x } _ { i } , j \right) }$} \\
			\multirow{3}{*}{} & \multirow{3}{*}{} \\
			\multirow{3}{*}{} & \multirow{3}{*}{} \\
			\multirow{3}{*}{\small   macro-AUC $\uparrow$}
			& \multirow{3}{*}{\small $\frac { 1 } { l } \sum \limits_ { j = 1 } ^ { l } \frac { \left| \mathcal { S } _ { \mathrm { macro } } ^ { j } \right| } { \left| Y _ { \cdot i } ^ { + } \right| \left| Y _ { \cdot i } ^ { - } \right| }$}
			\\
			\multirow{3}{*}{} & \multirow{3}{*}{} \\
			\multirow{3}{*}{} & \multirow{3}{*}{} \\
			\midrule
			\multicolumn{2}{l}{\small \multirow{2}{*}{$ \mathcal { S } _ { \mathrm { rank } } ^ { i } = \{ ( u , v ) \in Y _ { i } ^ { + } \times Y _ { i } ^ { - } | f _ { u } \left( \boldsymbol { x } _ { i } \right) \leq f _ { v } \left( \boldsymbol { x } _ { i } \right)$\}}} \\
			\multirow{2}{*}{} \\
			\multicolumn{2}{l}{\small \multirow{2}{*}{$ \mathcal { S } _ {\mathrm{precision}} ^ { i j } = \left\{ k \in Y _ { i } ^ { + } | \mathrm{rank} _ { F } \left( \boldsymbol { x } _ { i } , k \right) \leq \mathrm{rank} _ { F } \left( \boldsymbol { x } _ { i } , j \right) \right\}$}} \\
			\multirow{2}{*}{} \\
			\multicolumn{2}{l}{\small \multirow{2}{*}{$ \mathcal { S } _ { \mathrm { macro } } ^ { j } = \left\{ ( a , b ) \in Y _ { \cdot j } ^ { + } \times Y _ { \cdot j } ^ { - } | f _ { j } \left( \boldsymbol { x } _ { a } \right) \geq f _ { j } \left( \boldsymbol { x } _ { b } \right) \right\}$}} \\
			\multirow{2}{*}{} \\
			\bottomrule
		\end{tabular}		
	\end{center}
\end{table}

\subsection{Tree-based multi-label methods}
\label{multi-label forest}

Tree-based multi-label methods, such as ML-C4.5 \citep{clare2001knowledge} and PCT \citep{blockeel2000top}, are adapted from multi-class tree methods.
ML-C4.5, adapted from C4.5, allows multiple labels in the leaves of the tree, whose formula of entropy calculation is modified by summing the entropy for each label.
Predictive Clustering Tree (PCT) recursively partitions all samples into smaller clusters while moving down the tree.
In the testing process, the leaf node of a multi-label tree returns a vector of probabilities that a sample belongs to each class, by counting the percentage of different classes of training examples at the leaf node where concerned instance falls.
Finally, the information kept in each leaf node is the probability that the instance owns each label.

The ability of a single tree is limited, but ensemble of the trees will greatly improve performance.
Random Forest of Predictive Clustering Trees (RF-PCT) \citep{kocev2013tree} and Random Forest of ML-C4.5 (RFML-C4.5) \citep{experiment} are ensembles that use PCT and ML-C4.5 as base classifiers respectively.
The same as random forest, these forests use bagging and choose different feature sets to obtain the diversity among the base classifiers. 
Given a test instance, the forest will produce an estimate of label distribution by averaging results across all trees.

\section{THE PROPOSED METHOD}
\label{method}
In this section, we propose a deep forest method for multi-label learning. 
Firstly, we introduce the framework of Multi-Label Deep Forest (MLDF).
Then, we discuss two proposed mechanisms: measure-aware feature reuse and measure-aware layer growth.

\begin{figure*}[tb]
    \begin{center}
        \centerline{\includegraphics[width=\linewidth]{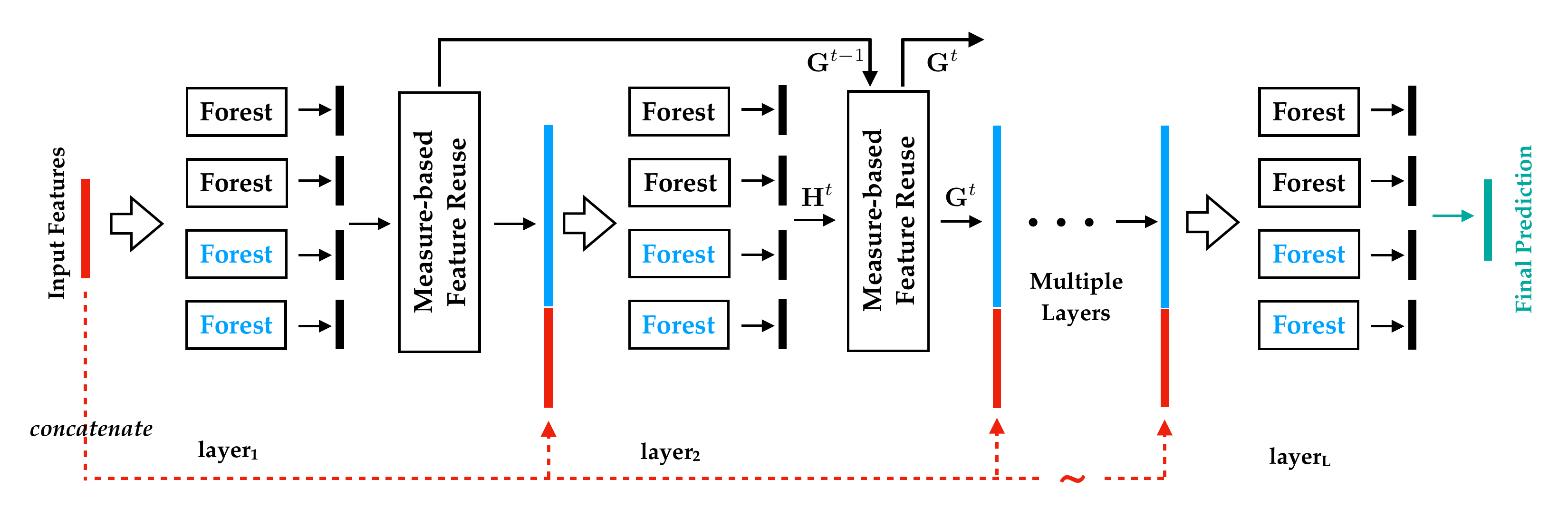}}
        \caption{The framework of Multi-Label Deep Forest (MLDF). Each layer ensembles two different forests (the black above and the blue below).}
        \label{fig:cascade}
    \end{center}
\end{figure*}

\subsection{The framework}
Figure \ref{fig:cascade} illustrates the framework of MLDF.
Different multi-label forests (the black forests above and the blue forests below) are ensembled in each layer of MLDF.
From $\mathrm{layer}_t$, we can obtain the representation $\mathbf{H}^t$.
The part of measure-aware feature reuse will receive the representation $\mathbf{H}^t$ and update it by reusing the representation $\mathbf{G}^{t-1}$ learned in the $\mathrm{layer}_{t-1}$ under the guidance of the performance of different measures.
Then the new representation $\mathbf{G}^{t}$ (the blue one) will be concatenated together with the raw input features (the red one) and goes into the next layer.

In MLDF, each layer is an ensemble of forests.
To enhance the performance of the ensemble, we consider different growing methods of trees to encourage diversity, which is crucial in the success of ensemble methods \citep{zhou2012ensemble}.
In traditional multi-class problems, extremely-random trees \citep{Geurts2006Extremely}, which takes one split point of each feature randomly, are used in gcForest \citep{zhou2017deep}.
For multi-label learning problems, we can also adapt this kind of method by changing the approach of splitting nodes when generating trees.
In MLDF, we take RF-PCT \citep{kocev2013tree} as the forest block, and two different methods generating nodes in trees are used to forests: one considers all possible split point of each feature, which is RF-PCT, the other considers one split point randomly \citep{E-PCT}, we name this as ERF-PCT.
Of course, other multi-label tree methods can also be embedded into each layer such as RFML-C4.5, which will be discussed in Section \ref{experiments}.

As aforementioned in Section \ref{multi-label forest}, given an instance, the forests will produce an estimation of label distributions, which can be viewed as the confidence of the instance belonging to each label.
The representation learned in each layer will adopt measure-aware feature reuse and be input to the next layer with raw input features.
The real-valued representation with rich labeling information will be input to the next layer to facilitate MLDF to take better advantage of label correlations \citep{Belanger2017End}.

The predicting process can be summarized as follows.
As shown in Figure \ref{fig:cascade}, assume the forests have been fitted well.
Firstly, we pre-process instances to standard matrix $\mathbf{X}$.
Secondly, the instances matrix $\mathbf{X}$ passes the first layer, and we can get the representation $\mathbf{H}^1$.
By adopting measure-aware feature reuse, we can get $\mathbf{G}^1$.
Then we concatenate $\mathbf{G}^1$ with the raw input features $\mathbf{X}$ and put them into the next layer.
After multiple layers, we get the final perdition.

\begin{table}
	\caption{Confidence computing method on six multi-label measures. $\boldsymbol{p}_{i\cdot}$ or $\boldsymbol{p}_{\cdot j}$ is sorted in descending order.}
	\label{confidence}
	\renewcommand\arraystretch{0.7}	
	\begin{center}
		\begin{tabular}{p{3cm}l}
			\toprule
			Measure & Confidence \\
			\midrule
			\multirow{3}{*}{\small   hamming loss} 
			& \multirow{3}{*}{\small $\frac { 1 } { m } \sum \limits_{ i = 1 } ^ { m } p_{ij}\mathbb{I}[p_{ij}>0.5] + (1-p_{ij})\mathbb{I}[p_{ij}\le 0.5]$}
			\\
			\multirow{3}{*}{} & \multirow{3}{*}{} \\
			\multirow{3}{*}{} & \multirow{3}{*}{} \\
			\multirow{3}{*}{\small one-error} 
			& \multirow{3}{*}{\small $\max \limits_{i=1,\dots,m}{p_{ij}}$}
			\\
			\multirow{3}{*}{} & \multirow{3}{*}{} \\
			\multirow{3}{*}{} & \multirow{3}{*}{} \\
			\multirow{3}{*}{\small   coverage} 
			& \multirow{3}{*}{\small $1 - \frac{1}{l}\sum \limits_ { j = 0 } ^ { l } \left[j \cdot p_{ij} \prod \limits_ { k = j+1 } ^ { l } \left(1 - p_{ik}\right)\right]$}
			\\
			\multirow{3}{*}{} & \multirow{4}{*}{} \\
			\multirow{3}{*}{} & \multirow{4}{*}{} \\
			\multirow{3}{*}{\small   ranking loss}  
			& \multirow{3}{*}{\small $\sum \limits_ { j = 0 } ^ { l } \prod \limits_ { k = 1 } ^ { j } p_{ik} \prod \limits_ { k = j+1 } ^ { l } \left(1 - p_{ik}\right)$}
			\\
			\multirow{3}{*}{} & \multirow{3}{*}{} \\
			\multirow{3}{*}{} & \multirow{3}{*}{} \\
			\multirow{3}{*}{\small   average precision}
			& \multirow{3}{*}{\small $\sum \limits_ { j = 0 } ^ { l } \prod \limits_ { k = 1 } ^ { j } p_{ik} \prod \limits_ { k = j+1 } ^ { l } \left(1 - p_{ik}\right)$} \\
			\multirow{3}{*}{} & \multirow{3}{*}{} \\
			\multirow{3}{*}{} & \multirow{3}{*}{} \\
			\multirow{3}{*}{\small   macro-AUC}
			& \multirow{3}{*}{\small $\sum \limits_ { i = 0 } ^ { m } \prod \limits_ { k = 1 } ^ { i } p_{kj} \prod \limits_ { k = i+1 } ^ { m } \left(1 - p_{kj}\right)$}
			\\
			\multirow{3}{*}{} & \multirow{3}{*}{} \\
			\multirow{3}{*}{} & \multirow{3}{*}{} \\
			\bottomrule
		\end{tabular}	
	\end{center}
\end{table}

\begin{algorithm}[t]
	\caption{Measure-aware feature reuse}
	\label{alg:MFS}
	\begin{algorithmic}[1]
		\renewcommand{\algorithmicrequire}{\textbf{Input:}}
		\REQUIRE  measure $M$, forests' output $\mathbf{H}^t$, previous $\mathbf{G}^{t-1}$.
		\renewcommand{\algorithmicensure}{\textbf{Output:}}
		\ENSURE new representation $\mathbf{G}^t$.
		\renewcommand{\algorithmicrequire}{\textbf{Procedure:}}
		\REQUIRE ~\\
		\STATE Initialize matrix $\mathbf{G}^t =\mathbf{H}^t$.
		\IF{Measure $M$ is label-based}
		\FOR{ $j=1$ {\bfseries to} $l$ }
		\STATE compute confidence $ \alpha^t_j$  on $\mathbf{H}^t_{\cdot j}$
		\STATE Update $\mathbf{G}^t_{\cdot j}$ to $\mathbf{G}^{t-1}_{\cdot j}$ when $\alpha^t_j < \theta_t$.
		\ENDFOR
		\ENDIF
		\IF{Measure $M$ is instance-based}
		\FOR{ $i=1$ {\bfseries to} $m$ }
		\STATE compute confidence $ \alpha^t_i$  on $\mathbf{H}^t_{i\cdot }$
		\STATE Update $\mathbf{G}^t_{i\cdot}$ to $\mathbf{G}^{t-1}_{i\cdot}$ when $\alpha^t_i < \theta_t$.
		\ENDFOR
		\ENDIF
	\end{algorithmic}
\end{algorithm}

\begin{algorithm}[t]
	\caption{Determine threshold}
	\label{alg:DT}
	\begin{algorithmic}[1]
		\renewcommand{\algorithmicrequire}{\textbf{Input:}}
		\REQUIRE  measure $M$, forests' output $\mathbf{H}^t$, ground-truth $\mathbf{Y}$, previous performance on $\mathrm{layer}_{t-1}$. 
		\renewcommand{\algorithmicensure}{\textbf{Output:}}
		\ENSURE threshold $\theta_t$.
		\renewcommand{\algorithmicrequire}{\textbf{Procedure:}}
		\REQUIRE ~\\
		\STATE Initialize confidence set $\mathcal{S} = \emptyset$. \\
		\IF{Measure $M$ is label-based}
		\FOR{ $j=1$ {\bfseries to} $l$ }
		\STATE compute confidence $ \alpha^t_j$ on $\mathbf{H}^t_{\cdot j}$.
		\STATE compute measure $m^t_j$ on $(\mathbf{H}^t_{\cdot j}, \mathbf{Y}_{\cdot j})$.
		\STATE $\mathcal{S} = \mathcal{S} \cup \{\alpha^t_j\}$ when $m^t_j$ is worse than $m^{t-1}_j$. 
		\ENDFOR
		\ENDIF
		\IF{Measure $M$ is instance-based}
		\FOR{ $i=1$ {\bfseries to} $m$ }
		\STATE compute confidence $ \alpha^t_i$ on $\mathbf{H}^t_{i \cdot}$.
		\STATE compute measure $m^t_i$ on $(\mathbf{H}^t_{i \cdot}, \mathbf{Y}_{i \cdot})$.
		\STATE $\mathcal{S} = \mathcal{S} \cup \{\alpha^t_i\}$ when $m^t_j$ is worse than $m^{t-1}_j$.
		\ENDFOR
		\ENDIF
		\STATE $\theta_t$ = Compute threshold on $\mathcal{S}$.
	\end{algorithmic}
\end{algorithm}

\subsection{Measure-aware feature reuse}
The split criterion of PCT is not directly related to the performance measure, and the representation $\mathbf{H}^t$ generated by each layer is same though the measure is different.
Therefore we propose the measure-aware feature reuse mechanism to enhance the representation under the guidance of different measures.
The key idea of measure-aware feature reuse is to partially reuse the good representation in the previous layer on the current layer if the confidence on the current layer is lower than a tehreshold determined in training, which can make the measure performance better.
Therefore, the challenge lies in defining the confidence of specific multi-label measures on demand.
Inspire by the confidence computing method on the multi-class classification problem \citep{pang1501improving}, we design a confidence computing method on different multi-label measures.

Hamming loss cares the correctness of single bit, one-error cares the element closest to 1, and others cares the ranking permutation on each row or column.
Therefore, the crucial step is designing reasonable methods to compute the confidence for various measures.
Table \ref{confidence} summarizes the computing method by considering the real meaning of each measure.
Matrix $\mathbf{P}$ is the average of $\mathbf{H}^t$, and the element $p_{ij}$ represents $\mathrm{Pr}[\hat{y}_{ij}=1]$.
For convenience, we arrange the element of each row (column) of $\mathbf{P}$ in descending order when the measure is instance-based (label-based).
Specifically, for hamming loss, we compute the max confidence that the bit is positive or negative.
For example, the prediction vector $\mathbf{p}_{\cdot j} = \left[0.9, 0.6, 0.4, 0.3\right]$, thus the confidence is $\alpha_j = \frac{1}{4}(0.9 + 0.6 + 0.6 + 0.7) = 0.7$.
For ranking loss, we compute the probability that 
the ranking loss is zero, which means the positive labels are ahead of negative labels.
For example, if the prediction vector $\mathbf{p}_{i\cdot} = \left[0.9, 0.6, 0.4, 0.3\right]$, there will be 5 possible permutations of ground-truth leading to zero ranking loss: $\{0000,1000,1100,1110,1111\}$.
Therefore, we can get the confidence by summing the probabilities in these five cases.
The confidence of macro-AUC and average precision are defined in a similarity way.

Algorithm \ref{alg:MFS} summarizes the process of the measure-aware feature reuse.
Due to the diversity between label-based measures and instance-based measures \citep{limo}, we need to deal with them separately.
Specifically, the label-based measures compute the confidence on each column of $\mathbf{H}^t$ and the instance-based measures compute it on each row.
After the confidence computing, we fix the previous representation $\mathbf{G}^{t-1}$  when the confidence $\alpha^t$ is below the threshold, and update $\mathbf{G}^{t}$ with the fixed ones.

The whole process of measure-aware feature reuse does not rely on true labels.
We can judge the goodness of representation by a threshold determined in training process.
As Algorithm \ref{alg:DT} shows, we save the confidence $\alpha^t$ into set $\mathcal{S}$ if the evaluated performance measure goes worse at $\mathrm{layer}_t$.
Then, the threshold $\theta_t$ is determined based on $\mathcal{S}$.
Simply, we can take an average on $\mathcal{S}$.
Because the meaning of confidence is consistent with the measure, the threshold $\theta_t$ can be effectively utilized in measure-aware feature reuse.

\subsection{Measure-aware layer growth}
\label{train}

\begin{algorithm}[t]
	\caption{Measure-aware layer growth}
	\label{alg:MLDF}
	\begin{algorithmic}[1]
		\renewcommand{\algorithmicrequire}{\textbf{Input:}}
		\REQUIRE maximal depth $T$, measure $M$, training data $\{\bf{X},\bf{Y}\}$.
		\renewcommand{\algorithmicensure}{\textbf{Output:}}
		\ENSURE model set $S$, threshold set $\Theta$ and final layer index $L$.
		\renewcommand{\algorithmicrequire}{\textbf{Procedure:}}
		\REQUIRE ~\\
		\STATE Initialize parameters: \\
		\quad performance in each layer $\boldsymbol{p}$[1:T],\\
		\quad best performance on train set $p_{\mathrm{best}}$,\\
		\quad the initial threshold $\theta_1 = 0$,\\
		\quad the best performance layer index $L=1$,\\
		\quad the model set $\mathcal{S} = \emptyset$.
		\FOR{ $t=1$ {\bfseries to} $T$ } \label{start}
		\STATE Train forests in $\text{layer}_t$ and get classifier $h_t$.\label{crucial step}
		\STATE Predict $\mathbf{H}_t = h_t([\mathbf{X}, \mathbf{G}_{t-1}])$. \label{second_step}
		\STATE $\theta_t = $ Determine Threshold (Algorithm \ref{alg:DT}) when $t > 2$. \label{Threshold}
		\STATE $\mathbf{G}^t = $ measure-aware feature reuse (Algorithm \ref{alg:MFS}). \label{MFR_step}
		\STATE Compute performance $\boldsymbol{p}[t]$ on measure $M$ with $\mathbf{G}^t$.
		\IF{ $\boldsymbol{p}[t]$ is better than $p_{\mathrm{best}}$ }
		\STATE Update best performance $p_{\mathrm{best}} = \boldsymbol{p}[t]$.
		\STATE Update the layer index of best performance $L = t$.
		\ELSIF{ $p_{\mathrm{best}}$ is not updated in recent 3 layers } \label{stop}
		\STATE \textbf{break}
		\ENDIF
		\STATE Add $\mathrm{layer}_t$ to model set: $\mathcal{S} = \mathcal{S} \cup \mathrm{layer}_t$. \label{add_step}
		\ENDFOR
		\STATE Keep $\{\mathrm{layer}_1,\dots,\mathrm{layer}_L\}$ in model set $S$ and drop others $\{\mathrm{layer}_{L+1},\dots\}$.
	\end{algorithmic}
\end{algorithm}

Though the measure-aware feature reuse can effectively enhance the representation guided by various measures, the mechanism can not influence the layer growth and reduce the risk of overfitting that may occur in training process. 
In order to reduce overfitting and control the model complexity, we propose the measure-aware layer growth mechanism.

If we use the same data to fit forests and do predicting directly, the risk of overfitting will be increased \citep{platt1999probabilistic}.
MLDF uses the $k$-fold cross-validation to alleviate this issue.
For each fold, we train the forests based on the training examples in other folds and make the prediction on the current fold. The layer's representation is generated by concatenating the predictions from each forest.

MLDF is built layer by layer.
Algorithm \ref{alg:MLDF} summarizes the procedure of measure-aware layer growth used in training MLDF.
The inputs are maximal depth of layers $T$, evaluation measure $M$, and the training data $\{\bf{X},\bf{Y}\}$.
If the model has grown $T$ layers, the training process will stop.
In general, we can choose one RF-PCT and one ERF-PCT in each layer and randomly select $\sqrt{d}$ number of features as candidates in each forest.
All parameters of training forest in MLDF, such as the number of forests and the depth of trees, are pre-determined before training.
As we hope each layer to learn different representations, we can set the maximum depth of tree in forests growing with the layer $i$, so does the number of trees, which can be set in advance.
In the initialization step, the performance vector $\boldsymbol{p}$, which records the performance value on training data in each layer, should be initialized according to different measures.
During each layer, we first fit the forests (line \ref{crucial step}) and get the representation $\mathbf{H}^t$ (line \ref{second_step}).
Then we should determine the threshold $\theta_t$ (line \ref{Threshold}) and generate new representation by measure-aware feature reuse (line \ref{MFR_step}).
Finally we add the $\mathrm{layer}_t$ to model set $\mathcal{S}$ (line \ref{add_step}).

The layer growth is measure-aware.
After fitting one layer, we need to compute the evaluation measure.
When the measure is not getting better in recent three layers (line \ref{stop}), MLDF is forced to stop growing. 
At the same time, the layer index of best performance in training dataset should be recorded, which is useful in prediction.
According to Occam's razor rule, we prefer a simpler model when the performance is similar.
While there is no clear improvement in performance, 
the final model set $\mathcal{S}=\{\mathrm{layer}_1,\dots,\mathrm{layer}_L\}$ should be kept, and layers after will be dropped.

Different measures represent different user demands. 
We can set $M$ as the required measure according to different situations.
Therefore, we can obtain the corresponding model for the specified measure.
On one hand, the model complexity can be controlled by $M$. 
On the other hand, compared to other methods, like PCT,  that cannot explicitly take the performance measure into consideration in the training process, MLDF is more flexible and has the ability to get better performance.

\section{EXPERIMENTS}
\label{experiments}

In this section, we conduct experiments with MLDF on different multi-label classification benchmark datasets.
Our goal is to validate MLDF can achieve the best performance on different measures and the two measure-aware mechanisms are necessray. 
Moreover, we show the advantages of MLDF through more detailed experiments from various aspects.

\subsection{Dataset and configuration}
We choose 9 multi-label classification benchmark datasets from different application domains and with different scales.
Table \ref{table-dataset} presents the basic statistics of these datasets.
All datasets are from a repository of multi-label datasets\footnote{http://mulan.sourceforge.net/datasets-mlc.html}. 
The datasets vary in size: from 502 up to 43970 examples, from 68 up to 5000 features, and from 5 up to 201 labels.
They are roughly organized in ascending order of the number of examples $m$, with eight of them being regular-scale, i.e., $m<5000$ and eight of them being large-scale, i.e., $m>5000$.
For all experiments conducted below, 50\% examples are randomly sampled without replacement to form the training set, and the rest 50\% examples are used to create the test set.

Six evaluation measures, widely used in multi-label learning \citep{limo}, are employed in this paper: hamming loss, one-error, coverage, ranking loss, average precision, and macro-AUC.
Note that the coverage is normalized by the number of labels and thus all the evaluation measures all vary between [0,1].

Hyper-parameters of MLDF is set as follows. We set the number of max layer ($T$) as 20 and take $M$ with the six measures discussed above respectively, which means we will get different models with different measures.
We take one RF-PCT and one ERF-PCT in each layer and use the 5-fold cross-validation to avoid overfitting.
In the first layer, we take 40 trees in each forest, and then take 20 more trees than the previous layer until the number of trees reaches 100, which can enable MLDF to learn different representations.
Similarly, we set the max-depth as 3 in the first layer, and then take 3 more than the previous layer when layer increasing.

\begin{table}
	\caption{Descriptions of the datasets in terms of the domain (Domain), number of examples ($m$), number of features ($d$) and number of labels ($l$).}
	\label{table-dataset}
	\begin{center}
		\begin{tabular}{p{2cm}p{2cm}p{1.5cm}p{1.5cm}p{1.5cm}}
			\toprule
			Dataset & Domain & $m$ & $d$  &$l$ \\
			\midrule
			CAL500   	& music 	 & 502  & 68   & 174   \\
			enron    	& text  	 & 1702 & 1001 & 53    \\
			image 	    & images 	 & 2000 & 294  &5	   \\
			scene    	& images 	 & 2407 & 1196 & 6     \\
			yeast    	& biology    & 2417 & 103  & 14    \\
			corel16k-s1 & images 	 & 13766& 500  & 153   \\
			corel16k-s2 & images 	 & 13761& 500  & 164   \\
			eurlex-sm   & text       & 19348& 5000 &201    \\
			mediamill   & multimedia & 43970& 120  & 101   \\
			\bottomrule
		\end{tabular}
	\end{center}
\end{table}

\begin{table*}[!thb] 
\caption{ Predictive performance (mean $\pm$ standard deviation) of each comparing methods on the nine datasets. $\bullet(\circ)$ indicates MLDF is significantly better(worse) than the comparing method on the criterion based on paired $t$-test at 95\% significant level. $\downarrow(\uparrow)$ means the smaller (larger) the value is, the performance will be the better.}
\label{regular}
\begin{scriptsize}
\begin{center}
\setlength{\tabcolsep}{0.4mm}{\begin{tabular}{p{1cm}ccccccccc}
\toprule 
\multicolumn{1}{l}{\multirow{1}{*}{Algorithm}} & \multicolumn{1}{c}{CAL500}  & \multicolumn{1}{c}{enron}  & \multicolumn{1}{c}{image}  & \multicolumn{1}{c}{scene}  & \multicolumn{1}{c}{yeast}  & \multicolumn{1}{c}{corel16k-s1} & \multicolumn{1}{c}{corel16k-s2} & \multicolumn{1}{c}{eurlex-sm} & \multicolumn{1}{c}{mediamill}\\
\midrule 
\multicolumn{8}{l}{hamming loss $\downarrow$} \\
\midrule 
MLDF & \multicolumn{1}{l}{0.136$\pm$0.001}  & \multicolumn{1}{l}{0.046$\pm$0.000}  & \multicolumn{1}{l}{0.148$\pm$0.003}  & \multicolumn{1}{l}{0.082$\pm$0.002}  & \multicolumn{1}{l}{0.190$\pm$0.003}  & \multicolumn{1}{l}{0.018$\pm$0.000}  & \multicolumn{1}{l}{0.017$\pm$0.000}  & \multicolumn{1}{l}{0.006$\pm$0.001}  & \multicolumn{1}{l}{0.027$\pm$0.001} \\
RF-PCT & \multicolumn{1}{l}{0.137$\pm$0.001$ \bullet$}  & \multicolumn{1}{l}{0.049$\pm$0.001$ \bullet$}  & \multicolumn{1}{l}{0.156$\pm$0.002$ \bullet$}  & \multicolumn{1}{l}{0.096$\pm$0.001$ \bullet$}  & \multicolumn{1}{l}{0.196$\pm$0.002$ \bullet$}  & \multicolumn{1}{l}{0.019$\pm$0.001$ \bullet$}  & \multicolumn{1}{l}{0.018$\pm$0.001$ \bullet$}  & \multicolumn{1}{l}{0.008$\pm$0.001$ \bullet$}  & \multicolumn{1}{l}{0.028$\pm$0.001$ \bullet$} \\
DBPNN & \multicolumn{1}{l}{0.169$\pm$0.001$ \bullet$}  & \multicolumn{1}{l}{0.075$\pm$0.001$ \bullet$}  & \multicolumn{1}{l}{0.264$\pm$0.001$ \bullet$}  & \multicolumn{1}{l}{0.260$\pm$0.001$ \bullet$}  & \multicolumn{1}{l}{0.220$\pm$0.001$ \bullet$}  & \multicolumn{1}{l}{0.029$\pm$0.001$ \bullet$}  & \multicolumn{1}{l}{0.026$\pm$0.001$ \bullet$}  & \multicolumn{1}{l}{0.007$\pm$0.001$ \bullet$}  & \multicolumn{1}{l}{0.031$\pm$0.001$ \bullet$} \\
MLFE & \multicolumn{1}{l}{0.141$\pm$0.002$ \bullet$}  & \multicolumn{1}{l}{0.047$\pm$0.001$ \bullet$}  & \multicolumn{1}{l}{0.162$\pm$0.006$ \bullet$}  & \multicolumn{1}{l}{0.084$\pm$0.002$ \bullet$}  & \multicolumn{1}{l}{0.203$\pm$0.002$ \bullet$}  & \multicolumn{1}{l}{0.019$\pm$0.001$ \bullet$}  & \multicolumn{1}{l}{0.018$\pm$0.001$ \bullet$}  & \multicolumn{1}{l}{0.007$\pm$0.001$ \bullet$}  & \multicolumn{1}{l}{0.029$\pm$0.001$ \bullet$} \\
ECC & \multicolumn{1}{l}{0.182$\pm$0.005$ \bullet$}  & \multicolumn{1}{l}{0.056$\pm$0.001$ \bullet$}  & \multicolumn{1}{l}{0.218$\pm$0.027$ \bullet$}  & \multicolumn{1}{l}{0.096$\pm$0.003$ \bullet$}  & \multicolumn{1}{l}{0.207$\pm$0.003$ \bullet$}  & \multicolumn{1}{l}{0.030$\pm$0.001$ \bullet$}  & \multicolumn{1}{l}{0.018$\pm$0.001$ \bullet$}  & \multicolumn{1}{l}{0.010$\pm$0.001$ \bullet$}  & \multicolumn{1}{l}{0.035$\pm$0.001$ \bullet$} \\
RAKEL & \multicolumn{1}{l}{0.138$\pm$0.002$ \bullet$}  & \multicolumn{1}{l}{0.058$\pm$0.001$ \bullet$}  & \multicolumn{1}{l}{0.173$\pm$0.004$ \bullet$}  & \multicolumn{1}{l}{0.096$\pm$0.004$ \bullet$}  & \multicolumn{1}{l}{0.202$\pm$0.003$ \bullet$}  & \multicolumn{1}{l}{0.020$\pm$0.001$ \bullet$}  & \multicolumn{1}{l}{0.019$\pm$0.001$ \bullet$}  & \multicolumn{1}{l}{0.007$\pm$0.001$ \bullet$}  & \multicolumn{1}{l}{0.031$\pm$0.001$ \bullet$} \\
\midrule 
\multicolumn{8}{l}{one-error $\downarrow$} \\
\midrule 
MLDF & \multicolumn{1}{l}{0.122$\pm$0.009}  & \multicolumn{1}{l}{0.216$\pm$0.009}  & \multicolumn{1}{l}{0.239$\pm$0.008}  & \multicolumn{1}{l}{0.188$\pm$0.005}  & \multicolumn{1}{l}{0.223$\pm$0.010}  & \multicolumn{1}{l}{0.640$\pm$0.003}  & \multicolumn{1}{l}{0.639$\pm$0.004}  & \multicolumn{1}{l}{0.138$\pm$0.001}  & \multicolumn{1}{l}{0.147$\pm$0.005} \\
RF-PCT & \multicolumn{1}{l}{0.122$\pm$0.010$ \bullet$}  & \multicolumn{1}{l}{0.231$\pm$0.011$ \bullet$}  & \multicolumn{1}{l}{0.258$\pm$0.005$ \bullet$}  & \multicolumn{1}{l}{0.215$\pm$0.010$ \bullet$}  & \multicolumn{1}{l}{0.247$\pm$0.008$ \bullet$}  & \multicolumn{1}{l}{0.723$\pm$0.002$ \bullet$}  & \multicolumn{1}{l}{0.721$\pm$0.006$ \bullet$}  & \multicolumn{1}{l}{0.270$\pm$0.006$ \bullet$}  & \multicolumn{1}{l}{0.150$\pm$0.002$ \bullet$} \\
DBPNN & \multicolumn{1}{l}{0.116$\pm$0.013$ \bullet$}  & \multicolumn{1}{l}{0.490$\pm$0.012$ \bullet$}  & \multicolumn{1}{l}{0.505$\pm$0.012$ \bullet$}  & \multicolumn{1}{l}{0.690$\pm$0.003$ \bullet$}  & \multicolumn{1}{l}{0.247$\pm$0.004$ \bullet$}  & \multicolumn{1}{l}{0.740$\pm$0.004$ \circ$}  & \multicolumn{1}{l}{0.697$\pm$0.004$ \bullet$}  & \multicolumn{1}{l}{0.460$\pm$0.015$ \bullet$}  & \multicolumn{1}{l}{0.200$\pm$0.003$ \bullet$} \\
MLFE & \multicolumn{1}{l}{0.133$\pm$0.010$ \bullet$}  & \multicolumn{1}{l}{0.232$\pm$0.003$ \bullet$}  & \multicolumn{1}{l}{0.265$\pm$0.008$ \circ$}  & \multicolumn{1}{l}{0.201$\pm$0.005$ \bullet$}  & \multicolumn{1}{l}{0.245$\pm$0.010$ \bullet$}  & \multicolumn{1}{l}{0.680$\pm$0.005$ \bullet$}  & \multicolumn{1}{l}{0.665$\pm$0.004$ \bullet$}  & \multicolumn{1}{l}{0.345$\pm$0.010$ \bullet$}  & \multicolumn{1}{l}{0.151$\pm$0.002$ \bullet$} \\
ECC & \multicolumn{1}{l}{0.137$\pm$0.021$ \bullet$}  & \multicolumn{1}{l}{0.293$\pm$0.008$ \bullet$}  & \multicolumn{1}{l}{0.408$\pm$0.069$ \bullet$}  & \multicolumn{1}{l}{0.247$\pm$0.010$ \bullet$}  & \multicolumn{1}{l}{0.244$\pm$0.009$ \bullet$}  & \multicolumn{1}{l}{0.706$\pm$0.006$ \bullet$}  & \multicolumn{1}{l}{0.712$\pm$0.005$ \bullet$}  & \multicolumn{1}{l}{0.346$\pm$0.007$ \bullet$}  & \multicolumn{1}{l}{0.150$\pm$0.005$ \bullet$} \\
RAKEL & \multicolumn{1}{l}{0.286$\pm$0.039$ \bullet$}  & \multicolumn{1}{l}{0.412$\pm$0.016$ \bullet$}  & \multicolumn{1}{l}{0.312$\pm$0.010$ \bullet$}  & \multicolumn{1}{l}{0.247$\pm$0.009$ \bullet$}  & \multicolumn{1}{l}{0.251$\pm$0.008$ \bullet$}  & \multicolumn{1}{l}{0.886$\pm$0.007$ \bullet$}  & \multicolumn{1}{l}{0.897$\pm$0.006$ \bullet$}  & \multicolumn{1}{l}{0.447$\pm$0.016$ \bullet$}  & \multicolumn{1}{l}{0.181$\pm$0.002$ \bullet$} \\
\midrule 
\multicolumn{8}{l}{coverage $\downarrow$} \\ 
\midrule 
MLDF & \multicolumn{1}{l}{0.274$\pm$0.005}  & \multicolumn{1}{l}{0.128$\pm$0.001}  & \multicolumn{1}{l}{0.159$\pm$0.004}  & \multicolumn{1}{l}{0.741$\pm$0.006}  & \multicolumn{1}{l}{0.434$\pm$0.004}  & \multicolumn{1}{l}{0.064$\pm$0.003}  & \multicolumn{1}{l}{0.262$\pm$0.002}  & \multicolumn{1}{l}{0.223$\pm$0.003}  & \multicolumn{1}{l}{0.066$\pm$0.001} \\
RF-PCT & \multicolumn{1}{l}{0.310$\pm$0.002$ \bullet$}  & \multicolumn{1}{l}{0.133$\pm$0.001$ \bullet$}  & \multicolumn{1}{l}{0.170$\pm$0.004$ \bullet$}  & \multicolumn{1}{l}{0.756$\pm$0.007$ \bullet$}  & \multicolumn{1}{l}{0.436$\pm$0.007}  & \multicolumn{1}{l}{0.073$\pm$0.004$ \bullet$}  & \multicolumn{1}{l}{0.321$\pm$0.002$ \bullet$}  & \multicolumn{1}{l}{0.223$\pm$0.007}  & \multicolumn{1}{l}{0.058$\pm$0.001$ \circ$} \\
DBPNN & \multicolumn{1}{l}{0.372$\pm$0.002$ \bullet$}  & \multicolumn{1}{l}{0.575$\pm$0.003$ \bullet$}  & \multicolumn{1}{l}{0.187$\pm$0.006$ \bullet$}  & \multicolumn{1}{l}{0.784$\pm$0.002$ \bullet$}  & \multicolumn{1}{l}{0.458$\pm$0.003$ \bullet$}  & \multicolumn{1}{l}{0.084$\pm$0.004$ \bullet$}  & \multicolumn{1}{l}{0.370$\pm$0.002$ \bullet$}  & \multicolumn{1}{l}{0.292$\pm$0.006$ \bullet$}  & \multicolumn{1}{l}{0.552$\pm$0.011$ \bullet$} \\
MLFE & \multicolumn{1}{l}{0.366$\pm$0.001$ \bullet$}  & \multicolumn{1}{l}{0.172$\pm$0.001$ \bullet$}  & \multicolumn{1}{l}{0.168$\pm$0.006$ \bullet$}  & \multicolumn{1}{l}{0.758$\pm$0.008$ \bullet$}  & \multicolumn{1}{l}{0.461$\pm$0.008$ \bullet$}  & \multicolumn{1}{l}{0.080$\pm$0.008$ \bullet$}  & \multicolumn{1}{l}{0.368$\pm$0.002$ \bullet$}  & \multicolumn{1}{l}{0.237$\pm$0.007$ \bullet$}  & \multicolumn{1}{l}{0.085$\pm$0.002$ \bullet$} \\
ECC & \multicolumn{1}{l}{0.436$\pm$0.002$ \bullet$}  & \multicolumn{1}{l}{0.467$\pm$0.009$ \bullet$}  & \multicolumn{1}{l}{0.229$\pm$0.034$ \bullet$}  & \multicolumn{1}{l}{0.806$\pm$0.016$ \bullet$}  & \multicolumn{1}{l}{0.464$\pm$0.005$ \bullet$}  & \multicolumn{1}{l}{0.084$\pm$0.002$ \bullet$}  & \multicolumn{1}{l}{0.446$\pm$0.003$ \bullet$}  & \multicolumn{1}{l}{0.349$\pm$0.014$ \bullet$}  & \multicolumn{1}{l}{0.386$\pm$0.010$ \bullet$} \\
RAKEL & \multicolumn{1}{l}{0.666$\pm$0.001$ \bullet$}  & \multicolumn{1}{l}{0.560$\pm$0.002$ \bullet$}  & \multicolumn{1}{l}{0.209$\pm$0.009$ \bullet$}  & \multicolumn{1}{l}{0.971$\pm$0.001$ \bullet$}  & \multicolumn{1}{l}{0.558$\pm$0.006$ \bullet$}  & \multicolumn{1}{l}{0.104$\pm$0.003$ \bullet$}  & \multicolumn{1}{l}{0.667$\pm$0.002$ \bullet$}  & \multicolumn{1}{l}{0.523$\pm$0.008$ \bullet$}  & \multicolumn{1}{l}{0.543$\pm$0.012$ \bullet$} \\
\midrule 
\multicolumn{8}{l}{ranking loss $\downarrow$} \\
\midrule 
MLDF & \multicolumn{1}{l}{0.176$\pm$0.002}  & \multicolumn{1}{l}{0.077$\pm$0.001}  & \multicolumn{1}{l}{0.129$\pm$0.005}  & \multicolumn{1}{l}{0.059$\pm$0.004}  & \multicolumn{1}{l}{0.160$\pm$0.006}  & \multicolumn{1}{l}{0.143$\pm$0.002}  & \multicolumn{1}{l}{0.138$\pm$0.002}  & \multicolumn{1}{l}{0.014$\pm$0.001}  & \multicolumn{1}{l}{0.034$\pm$0.001} \\
RF-PCT & \multicolumn{1}{l}{0.178$\pm$0.002$ \bullet$}  & \multicolumn{1}{l}{0.079$\pm$0.001$ \bullet$}  & \multicolumn{1}{l}{0.142$\pm$0.004$ \bullet$}  & \multicolumn{1}{l}{0.070$\pm$0.004$ \bullet$}  & \multicolumn{1}{l}{0.164$\pm$0.008$ \bullet$}  & \multicolumn{1}{l}{0.165$\pm$0.001$ \bullet$}  & \multicolumn{1}{l}{0.142$\pm$0.001$ \bullet$}  & \multicolumn{1}{l}{0.029$\pm$0.001$ \bullet$}  & \multicolumn{1}{l}{0.035$\pm$0.001$ \bullet$} \\
DBPNN & \multicolumn{1}{l}{0.185$\pm$0.002$ \bullet$}  & \multicolumn{1}{l}{0.126$\pm$0.007$ \bullet$}  & \multicolumn{1}{l}{0.278$\pm$0.005$ \bullet$}  & \multicolumn{1}{l}{0.277$\pm$0.005$ \bullet$}  & \multicolumn{1}{l}{0.187$\pm$0.001$ \bullet$}  & \multicolumn{1}{l}{0.154$\pm$0.002$ \bullet$}  & \multicolumn{1}{l}{0.148$\pm$0.002$ \bullet$}  & \multicolumn{1}{l}{0.396$\pm$0.011$ \bullet$}  & \multicolumn{1}{l}{0.230$\pm$0.001$ \bullet$} \\
MLFE & \multicolumn{1}{l}{0.185$\pm$0.003$ \bullet$}  & \multicolumn{1}{l}{0.082$\pm$0.008$ \bullet$}  & \multicolumn{1}{l}{0.148$\pm$0.007$ \bullet$}  & \multicolumn{1}{l}{0.065$\pm$0.004$ \bullet$}  & \multicolumn{1}{l}{0.174$\pm$0.006$ \bullet$}  & \multicolumn{1}{l}{0.189$\pm$0.002$ \bullet$}  & \multicolumn{1}{l}{0.188$\pm$0.001$ \bullet$}  & \multicolumn{1}{l}{0.034$\pm$0.002$ \bullet$}  & \multicolumn{1}{l}{0.046$\pm$0.001$ \bullet$} \\
ECC & \multicolumn{1}{l}{0.204$\pm$0.008$ \bullet$}  & \multicolumn{1}{l}{0.133$\pm$0.004$ \bullet$}  & \multicolumn{1}{l}{0.224$\pm$0.043$ \bullet$}  & \multicolumn{1}{l}{0.085$\pm$0.003$ \bullet$}  & \multicolumn{1}{l}{0.186$\pm$0.003$ \bullet$}  & \multicolumn{1}{l}{0.233$\pm$0.002$ \bullet$}  & \multicolumn{1}{l}{0.229$\pm$0.001$ \bullet$}  & \multicolumn{1}{l}{0.263$\pm$0.007$ \bullet$}  & \multicolumn{1}{l}{0.179$\pm$0.008$ \bullet$} \\
RAKEL & \multicolumn{1}{l}{0.444$\pm$0.005$ \bullet$}  & \multicolumn{1}{l}{0.241$\pm$0.005$ \bullet$}  & \multicolumn{1}{l}{0.196$\pm$0.008$ \bullet$}  & \multicolumn{1}{l}{0.107$\pm$0.003$ \bullet$}  & \multicolumn{1}{l}{0.245$\pm$0.004$ \bullet$}  & \multicolumn{1}{l}{0.414$\pm$0.002$ \bullet$}  & \multicolumn{1}{l}{0.418$\pm$0.001$ \bullet$}  & \multicolumn{1}{l}{0.388$\pm$0.011$ \bullet$}  & \multicolumn{1}{l}{0.222$\pm$0.001$ \bullet$} \\
\midrule 
\multicolumn{8}{l}{average precision $\uparrow$} \\ 
\midrule 
MLDF & \multicolumn{1}{l}{0.512$\pm$0.003}  & \multicolumn{1}{l}{0.696$\pm$0.004}  & \multicolumn{1}{l}{0.842$\pm$0.005}  & \multicolumn{1}{l}{0.891$\pm$0.008}  & \multicolumn{1}{l}{0.770$\pm$0.005}  & \multicolumn{1}{l}{0.347$\pm$0.002}  & \multicolumn{1}{l}{0.342$\pm$0.004}  & \multicolumn{1}{l}{0.840$\pm$0.002}  & \multicolumn{1}{l}{0.732$\pm$0.007} \\
RF-PCT & \multicolumn{1}{l}{0.512$\pm$0.006$ \bullet$}  & \multicolumn{1}{l}{0.685$\pm$0.002$ \bullet$}  & \multicolumn{1}{l}{0.829$\pm$0.003$ \bullet$}  & \multicolumn{1}{l}{0.873$\pm$0.006$ \bullet$}  & \multicolumn{1}{l}{0.758$\pm$0.008$ \bullet$}  & \multicolumn{1}{l}{0.293$\pm$0.002$ \bullet$}  & \multicolumn{1}{l}{0.287$\pm$0.002$ \bullet$}  & \multicolumn{1}{l}{0.726$\pm$0.004$ \bullet$}  & \multicolumn{1}{l}{0.729$\pm$0.001$ \bullet$} \\
DBPNN & \multicolumn{1}{l}{0.495$\pm$0.002$ \bullet$}  & \multicolumn{1}{l}{0.500$\pm$0.007$ \bullet$}  & \multicolumn{1}{l}{0.672$\pm$0.006$ \bullet$}  & \multicolumn{1}{l}{0.563$\pm$0.004$ \bullet$}  & \multicolumn{1}{l}{0.738$\pm$0.002$ \bullet$}  & \multicolumn{1}{l}{0.289$\pm$0.002$ \bullet$}  & \multicolumn{1}{l}{0.299$\pm$0.002$ \bullet$}  & \multicolumn{1}{l}{0.427$\pm$0.013$ \bullet$}  & \multicolumn{1}{l}{0.502$\pm$0.002$ \bullet$} \\
MLFE & \multicolumn{1}{l}{0.488$\pm$0.006$ \bullet$}  & \multicolumn{1}{l}{0.688$\pm$0.009$ \bullet$}  & \multicolumn{1}{l}{0.817$\pm$0.010$ \bullet$}  & \multicolumn{1}{l}{0.882$\pm$0.005$ \bullet$}  & \multicolumn{1}{l}{0.759$\pm$0.005$ \bullet$}  & \multicolumn{1}{l}{0.319$\pm$0.001$ \bullet$}  & \multicolumn{1}{l}{0.317$\pm$0.001$ \bullet$}  & \multicolumn{1}{l}{0.853$\pm$0.007$ \bullet$}  & \multicolumn{1}{l}{0.728$\pm$0.001$ \circ$} \\
ECC & \multicolumn{1}{l}{0.482$\pm$0.008$ \bullet$}  & \multicolumn{1}{l}{0.651$\pm$0.006$ \bullet$}  & \multicolumn{1}{l}{0.739$\pm$0.043$ \bullet$}  & \multicolumn{1}{l}{0.853$\pm$0.005$ \bullet$}  & \multicolumn{1}{l}{0.752$\pm$0.006$ \bullet$}  & \multicolumn{1}{l}{0.282$\pm$0.003$ \bullet$}  & \multicolumn{1}{l}{0.276$\pm$0.002$ \bullet$}  & \multicolumn{1}{l}{0.572$\pm$0.007$ \bullet$}  & \multicolumn{1}{l}{0.597$\pm$0.014$ \bullet$} \\
RAKEL & \multicolumn{1}{l}{0.353$\pm$0.006$ \bullet$}  & \multicolumn{1}{l}{0.539$\pm$0.006$ \bullet$}  & \multicolumn{1}{l}{0.788$\pm$0.006$ \bullet$}  & \multicolumn{1}{l}{0.843$\pm$0.005$ \bullet$}  & \multicolumn{1}{l}{0.720$\pm$0.005$ \bullet$}  & \multicolumn{1}{l}{0.103$\pm$0.003$ \bullet$}  & \multicolumn{1}{l}{0.092$\pm$0.003$ \bullet$}  & \multicolumn{1}{l}{0.440$\pm$0.013$ \bullet$}  & \multicolumn{1}{l}{0.521$\pm$0.001$ \bullet$} \\
\midrule 
\multicolumn{8}{l}{macro-AUC $\uparrow$} \\ 
\midrule 
MLDF & \multicolumn{1}{l}{0.568$\pm$0.006}  & \multicolumn{1}{l}{0.742$\pm$0.014}  & \multicolumn{1}{l}{0.885$\pm$0.003}  & \multicolumn{1}{l}{0.956$\pm$0.003}  & \multicolumn{1}{l}{0.732$\pm$0.010}  & \multicolumn{1}{l}{0.728$\pm$0.001}  & \multicolumn{1}{l}{0.737$\pm$0.007}  & \multicolumn{1}{l}{0.930$\pm$0.002}  & \multicolumn{1}{l}{0.842$\pm$0.002} \\
RF-PCT & \multicolumn{1}{l}{0.555$\pm$0.004$ \bullet$}  & \multicolumn{1}{l}{0.729$\pm$0.012$ \bullet$}  & \multicolumn{1}{l}{0.875$\pm$0.005$ \bullet$}  & \multicolumn{1}{l}{0.947$\pm$0.002$ \bullet$}  & \multicolumn{1}{l}{0.723$\pm$0.012$ \bullet$}  & \multicolumn{1}{l}{0.712$\pm$0.004$ \bullet$}  & \multicolumn{1}{l}{0.719$\pm$0.005$ \bullet$}  & \multicolumn{1}{l}{0.904$\pm$0.007$ \bullet$}  & \multicolumn{1}{l}{0.835$\pm$0.002$ \bullet$} \\
DBPNN & \multicolumn{1}{l}{0.499$\pm$0.001$ \bullet$}  & \multicolumn{1}{l}{0.679$\pm$0.010$ \bullet$}  & \multicolumn{1}{l}{0.746$\pm$0.006$ \bullet$}  & \multicolumn{1}{l}{0.704$\pm$0.005$ \bullet$}  & \multicolumn{1}{l}{0.627$\pm$0.004$ \bullet$}  & \multicolumn{1}{l}{0.699$\pm$0.002$ \bullet$}  & \multicolumn{1}{l}{0.708$\pm$0.003$ \bullet$}  & \multicolumn{1}{l}{0.589$\pm$0.005$ \bullet$}  & \multicolumn{1}{l}{0.510$\pm$0.001$ \bullet$} \\
MLFE & \multicolumn{1}{l}{0.547$\pm$0.006$ \bullet$}  & \multicolumn{1}{l}{0.656$\pm$0.010$ \bullet$}  & \multicolumn{1}{l}{0.841$\pm$0.006$ \bullet$}  & \multicolumn{1}{l}{0.944$\pm$0.004$ \bullet$}  & \multicolumn{1}{l}{0.705$\pm$0.005$ \bullet$}  & \multicolumn{1}{l}{0.651$\pm$0.006$ \bullet$}  & \multicolumn{1}{l}{0.662$\pm$0.002$ \bullet$}  & \multicolumn{1}{l}{0.853$\pm$0.003$ \bullet$}  & \multicolumn{1}{l}{0.799$\pm$0.002$ \bullet$} \\
ECC & \multicolumn{1}{l}{0.507$\pm$0.005$ \bullet$}  & \multicolumn{1}{l}{0.646$\pm$0.008$ \bullet$}  & \multicolumn{1}{l}{0.807$\pm$0.030$ \bullet$}  & \multicolumn{1}{l}{0.931$\pm$0.004$ \bullet$}  & \multicolumn{1}{l}{0.646$\pm$0.003$ \bullet$}  & \multicolumn{1}{l}{0.627$\pm$0.004$ \bullet$}  & \multicolumn{1}{l}{0.633$\pm$0.002$ \bullet$}  & \multicolumn{1}{l}{0.624$\pm$0.004$ \bullet$}  & \multicolumn{1}{l}{0.524$\pm$0.001$ \bullet$} \\
RAKEL & \multicolumn{1}{l}{0.547$\pm$0.007$ \bullet$}  & \multicolumn{1}{l}{0.596$\pm$0.007$ \bullet$}  & \multicolumn{1}{l}{0.803$\pm$0.005$ \bullet$}  & \multicolumn{1}{l}{0.884$\pm$0.004$ \bullet$}  & \multicolumn{1}{l}{0.614$\pm$0.003$ \bullet$}  & \multicolumn{1}{l}{0.523$\pm$0.001$ \bullet$}  & \multicolumn{1}{l}{0.525$\pm$0.001$ \bullet$}  & \multicolumn{1}{l}{0.591$\pm$0.006$ \bullet$}  & \multicolumn{1}{l}{0.513$\pm$0.001$ \bullet$} \\
\bottomrule 
\end{tabular}} 
\end{center}
\end{scriptsize}
\end{table*}

\subsection{Performance comparison}
\label{performance comparison}
We compare MLDF to the following five contenders: 
a) RF-PCT \citep{kocev2013tree}, 
b) DBPNN \citep{Hinton2006, MEKA},
c) MLFE \citep{ZhangZZ18}, 
d) RAKEL \citep{tsoumakas2007random} and
e) ECC \citep{ECC}.
In above, DBPNN is the representative of DNN methods; RAKEL, ECC, and RF-PCT are representatives of multi-label ensemble methods; MLFE is a method that utilizes the structural information in feature space to enrich the labeling information.
Parameter settings of the compared methods are listed below.
In details, for RF-PCT, we take the amount as 100.
For DBPNN, we conduct the experiments with Meka \citep{MEKA} and set the base classifier as the logistic function
, other hyper-parameters are the same as recommended in Meka
.
For MLFE, we keep the same setting as suggested in \citep{ZhangZZ18}, where $\rho =1, c_1 = 1, c_2 = 2$, $\beta_1$, $\beta_2$ and $\beta_3$ are chosen among $\{1,2,...,10\},\{1,10,15\}$ and $\{1,10\}$ respectively.
The ensemble size of ECC is set to 100 to accommodate the sufficient number of classifier chains, and the ensemble size of RAKEL is set to $2q$ with $k = 3$ as suggested in the literature.
The base learner of ECC and RAKEL is SVM with a linear kernel.
For fairness, all methods use the 5-fold cross-validation.

We conduct experiments on each algorithm for ten times.
The mean metric value and the standard deviation across 10 training/testing trials are recorded for comparative studies.
Table \ref{regular} reports the detailed experimental results of comparing algorithms.
MLDF achieves optimal (lowest) average rank in terms of each evaluation measure. On the 9 benchmark datasets, across all the evaluation measures, MLDF ranks 1st in 98.46\% cases and ranks 2nd in 1.54\% cases.
Compared on these six measures, MLDF ranks 1st in 100.00\%, 96.29\%, 96.29\%, 100.00\%, 98.15\%, 100.00\% cases respectively.
To summarize, MLDF achieves the best performance against other well-established contenders across extensive benchmark datasets on various evaluation measures, which validates the effectiveness of MLDF.

\subsection{Influence of measure-aware feature reuse}

\begin{figure}[b]
	\centering
	\includegraphics[width=\textwidth]{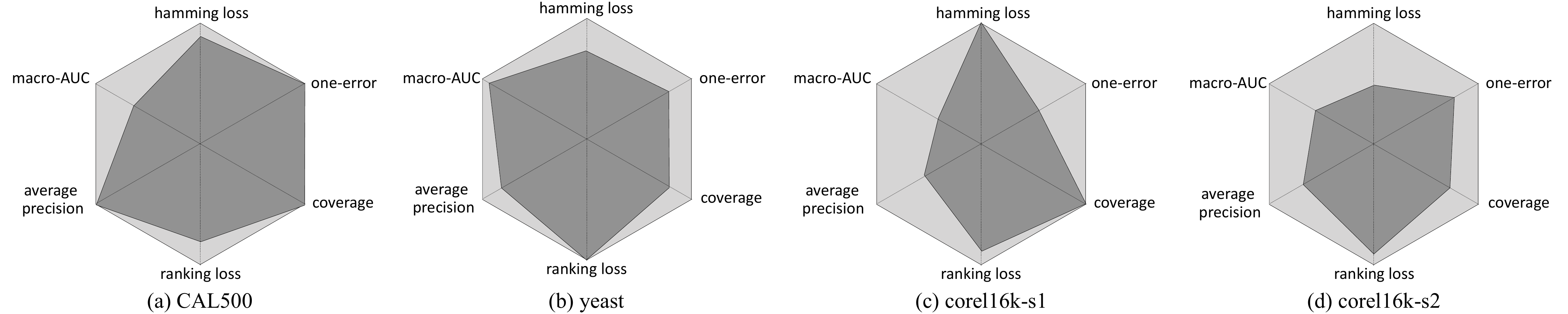}
	\caption{The performance comparison on CAL500, yeast, corel16k-s1 and corel16k-s2. The light hexagon represents the performance of MLDF, the darker one represents the performance of MLDF without the measure-aware feature reuse mechanism. The larger the area means the better performance.}
	\label{fig:skip}
\end{figure}

The measure-aware feature reuse aims to reuse the good representation in the previous layer according to confidence.
When the confidence $\alpha^t$ is lower than threshold $\theta_t$, we reuse the representation $\mathbf{G}^{t-1}$ in $\mathrm{layer}_{t-1}$.
If we skip the line \ref{Threshold} in Algorithm \ref{alg:MLDF} and keep $\theta_t$ as 0 for all t in $[1,L]$, it is just that we do not take the measure-aware feature reuse mechanism in all layers.
Figure \ref{fig:skip} shows the comparison between using the mechanism and not using the mechanism on CAL500, yeast, corel16k-s1 and corel16k-s2.
The six radii represent six different measures, the outermost part of the hexagon represents the performance of MLDF, the center represents the performance of RF-PCT, and the darker hexagon represents the performance of MLDF without the mechanism.
Area of MLDF is larger than that of MLDF without the mechanism, therefore it indicates that the measure-aware feature reuse mechanism is workable on these datasets.
Furthermore, the area of MLDF without the  measure-aware feature reuse mechanism gets smaller when the size of datasets increase, which confirms that the measure-aware feature reuse mechanism does well in larger data.

\subsection{Effect of measure-aware layer growth}

We conduct experiments on \emph{yeast} dataset to show effect of the measure-aware layer growth mechanism.
Specifically, when MLDF arrives its final layer index $L$, we keep the layer increase for observing whether the mechanism is effective.
The number of trees in RF-PCT is 100, and the 5-fold cross-validation is used.

\begin{figure}[tb]
	\begin{center}
		\centerline{\includegraphics[width=0.6\columnwidth]{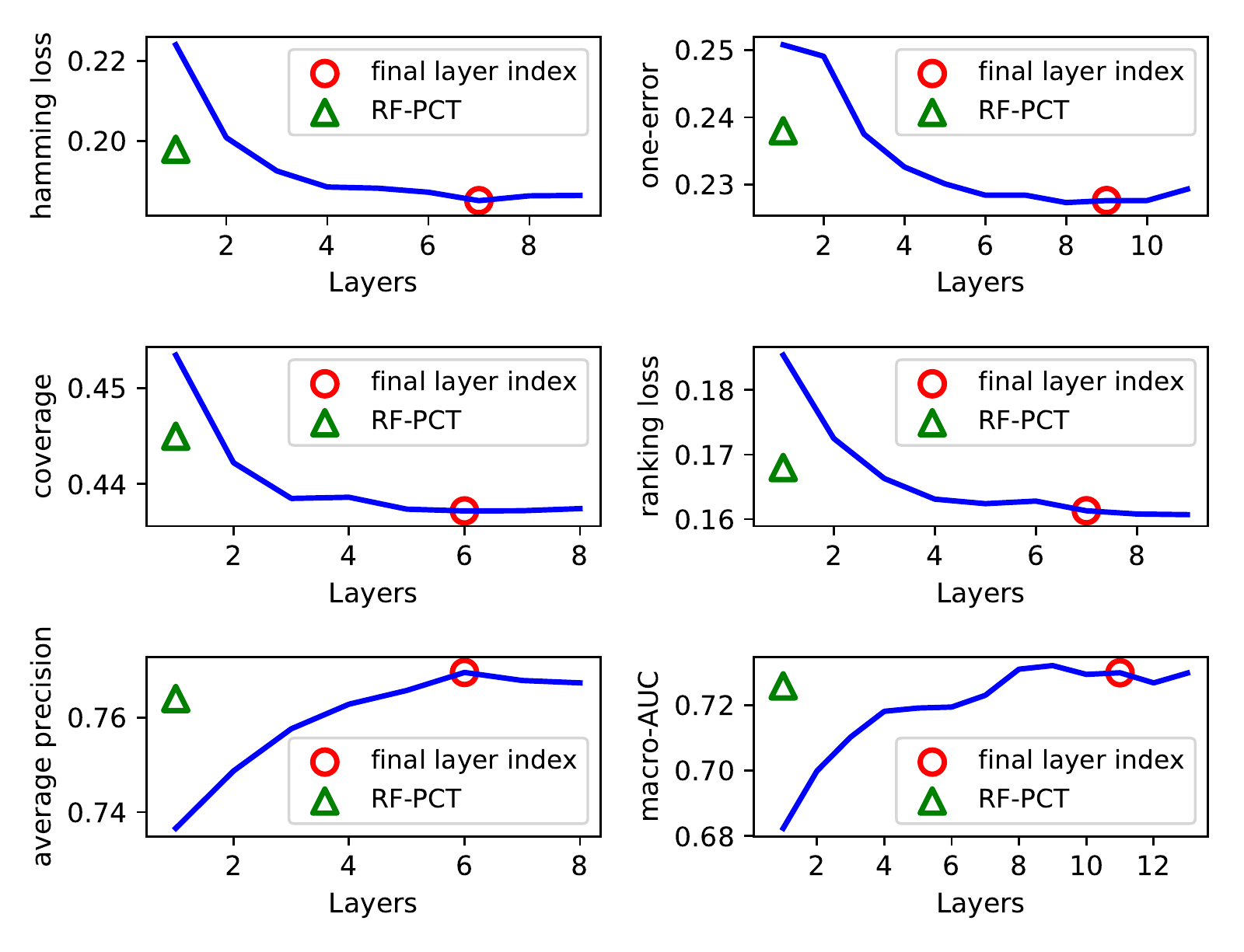}}
		\caption{The test performance of MLDF in each layer on six measures on \emph{yeast} dataset respectively. The triangle indicates the performance of RF-PCT, and the circle means the final layer index $L$.}
		\label{fig:depth}
	\end{center}
\end{figure}

Figure \ref{fig:depth} shows the \emph{yeast}'s test performance curve of MLDF on 6 measures.
The red circle means the final layer returned by Algorithm \ref{alg:MLDF}.
For RF-PCT, we set the same trees as the final layer of MLDF (the red circle).
The triangle indicates the performance of RF-PCT, which can be seen as a one-layer MLDF.
In Figure \ref{fig:depth}, the performance of MLDF becomes better when the model goes deeper, and our algorithm can stop almost at the position with the best performance.
It demonstrates the effectiveness of our stopping mechanism.
The performance of MLDF (the circle) is better than RF-PCT (the triangle), where they have the same amount of trees.
Moreover, MLDF controlled by different measures can converge in different layers.
It indicates that the measure-aware layer growth mechanism is effective.

\subsection{Label correlations exploitation}

\begin{table*}[tb] 
\caption{ Predictive performance (mean $\pm$ standard deviation) of each comparing methods on all datasets. ($\cdot/\cdot/\cdot$) indicates the times that MLDF is significantly (superior/ equal/ inferior) to the comparing method on the criterion based on paired $t$-test at 95\% significant level.}
\label{entropy}
\begin{scriptsize}
\begin{center}
\setlength{\tabcolsep}{0.5mm}{
\begin{tabular}{lcccccc} 
\toprule
dataset & hamming loss & one-error &coverage & ranking loss & average precision & macro-AUC \\ 
\midrule 
CAL500 & 0.137$\pm$0.001 (5/0/0) & 0.120$\pm$0.018 (4/0/1) & 0.738$\pm$0.004 (5/0/0) & 0.176$\pm$0.002 (5/0/0) & 0.511$\pm$0.005 (4/0/1) & 0.569$\pm$0.007 (5/0/0)\\ 
enron & 0.047$\pm$0.001 (5/0/0) & 0.225$\pm$0.007 (5/0/0) & 0.224$\pm$0.005 (4/0/1) & 0.079$\pm$0.004 (5/0/0) & 0.691$\pm$0.003 (5/0/0) & 0.738$\pm$0.005 (5/0/0)\\ 
image & 0.146$\pm$0.004 (5/0/0) & 0.243$\pm$0.016 (5/0/0) & 0.157$\pm$0.008 (5/0/0) & 0.130$\pm$0.010 (5/0/0) & 0.841$\pm$0.010 (5/0/0) & 0.886$\pm$0.010 (5/0/0)\\ 
scene & 0.083$\pm$0.003 (5/0/0) & 0.192$\pm$0.007 (5/0/0) & 0.063$\pm$0.002 (5/0/0) & 0.059$\pm$0.002 (5/0/0) & 0.889$\pm$0.004 (5/0/0) & 0.956$\pm$0.003 (5/0/0)\\ 
yeast & 0.190$\pm$0.003 (5/0/0) & 0.225$\pm$0.009 (5/0/0) & 0.434$\pm$0.006 (5/0/0) & 0.159$\pm$0.002 (5/0/0) & 0.769$\pm$0.003 (5/0/0) & 0.733$\pm$0.006 (5/0/0)\\ 
corel16k-s1 & 0.019$\pm$0.001 (5/0/0) & 0.726$\pm$0.002 (2/0/3) & 0.316$\pm$0.012 (5/0/0) & 0.163$\pm$0.007 (4/0/1) & 0.295$\pm$0.002 (4/0/1) & 0.695$\pm$0.002 (3/0/2)\\ 
corel16k-s2 & 0.018$\pm$0.001 (5/0/0) & 0.727$\pm$0.001 (1/0/4) & 0.313$\pm$0.016 (4/0/1) & 0.160$\pm$0.009 (3/0/2) & 0.284$\pm$0.007 (2/0/3) & 0.697$\pm$0.011 (3/0/2)\\ 
eurlex-sm & 0.007$\pm$0.001 (5/0/0) & 0.201$\pm$0.012 (5/0/0) & 0.043$\pm$0.001 (5/0/0) & 0.021$\pm$0.001 (5/0/0) & 0.784$\pm$0.008 (4/0/1) & 0.900$\pm$0.002 (4/0/1)\\ 
mediamill & 0.027$\pm$0.001 (5/0/0) & 0.144$\pm$0.002 (5/0/0) & 0.128$\pm$0.002 (5/0/0) & 0.035$\pm$0.001 (5/0/0) & 0.725$\pm$0.010 (3/0/2) & 0.843$\pm$0.004 (5/0/0)\\ 
\midrule 
sum of score & 45/0/0 & 37/0/8 & 43/0/2 & 42/0/3 & 37/0/8 & 40/0/5 \\ 
\bottomrule 
\end{tabular}} 
\end{center} 
\end{scriptsize} 
\end{table*}

\begin{figure*}[htb]
	\centerline{\includegraphics[width=0.6\textwidth]{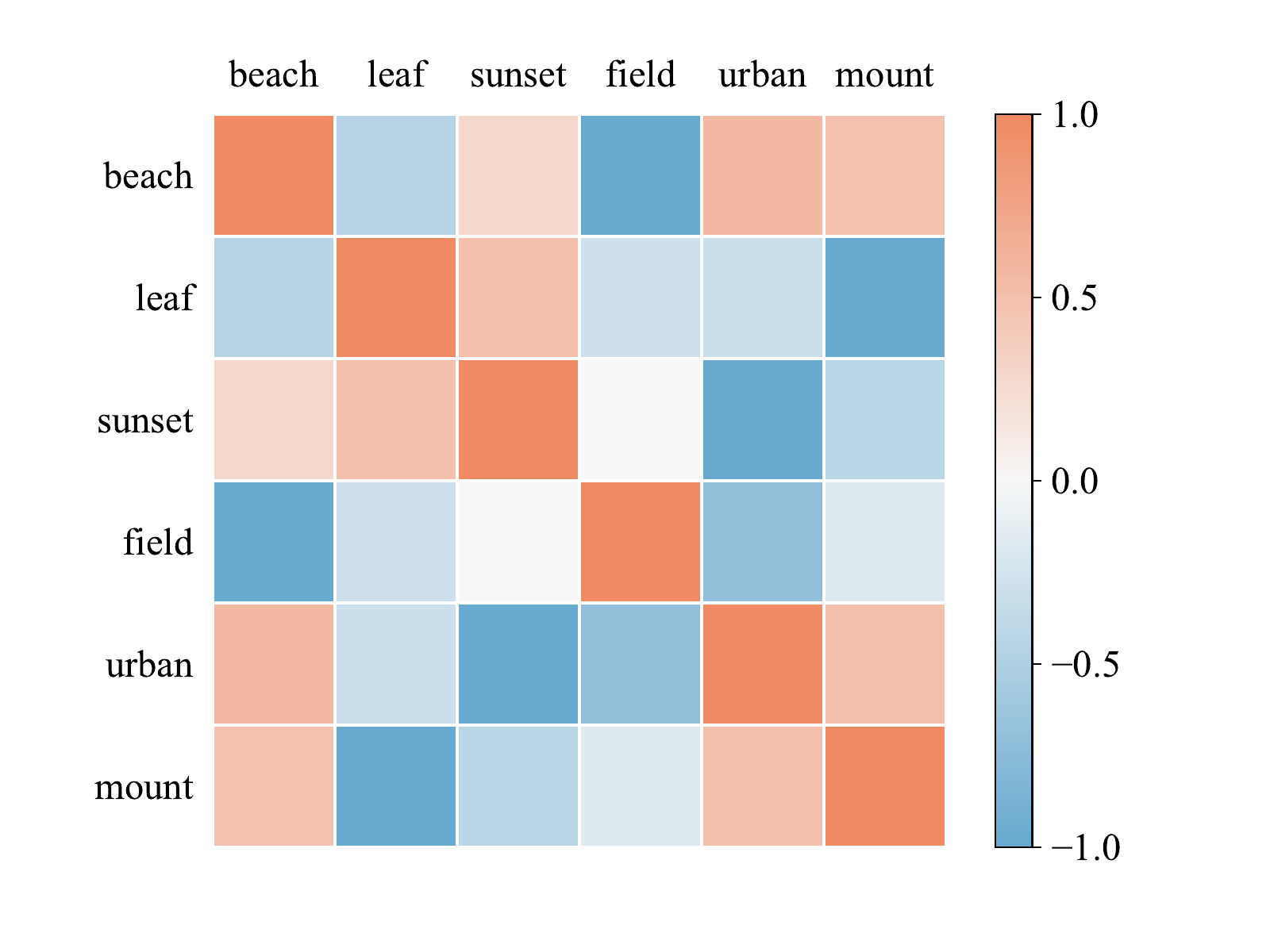}}
	\caption{Effect of missing representation information on each label respectively. The \emph{scene} dataset has 6 labels (top-to-down, left-to-right): ``Beach", ``Leaf", ``Sunset", ``Field", ``Urban" and ``Mountain". }
	\label{fig:correlation}
\end{figure*}

Intuitively, the cascade structure enables MLDF to utilize label correlations.
Thus we conduct a distinctive approach to exploit label correlations.
Our layer-wise method gradually considers more complex label correlations by utilizing the lower layer label representation in higher layer modelling. 
Here we deliberately delete a specific label in first layer representation, then train the second layer and check the influence of that malicious deletion. 
Suppose accuracy decrease on label B is observed after deleting label A, we consider the two labels are correlated. 
The relative normalized decrease indicates the strength of correlations and we show the result on \emph{scene} dataset in Figure \ref{fig:correlation}.
As shown in Figure \ref{fig:correlation}, label ``Sunset" highly effects label ``Leaf" each other, since missing one's information will effect the performance sharply on the other label.
Besides, label ``Beach" is highly correlated with label ``Urban'' since sometimes they exist in \emph{scene} dataset together \citep{boutell2004learning}.
It indicates that MLDF utilizes some correlations between labels to get better performance in inner layers.

\subsection{Flexibility}

In previous experiments, RF-PCT and ERF-PCT are the forest blocks in MLDF, which achieve the best performance.
A natural question will be \emph{is it possible to replace the forest block by other multi-label tree-based methods in MLDF, and how will the performance change}?
To investigate this problem, we take one RFML-C4.5 and one ERFML-C4.5 in MLDF.
To ensure fairness, we keep all the other configurations \emph{same} as those in Section \ref{performance comparison}.

Table \ref{entropy} shows the result of MLDF based on RFML-C4.5.
By comparing the results in Table \ref{regular}, we count the times that MLDF wins/draws/loses the comparison.
MLDF wins 100.0\%, 82.2\%, 95.6\%, 93.3\%, 82.2\%, 88.9\% among all the 45 comparisons on six evaluation measures respectively.
It is clear that MLDF based on RFML-C4.5 also achieves the best performance among all compared methods.
Thus, no matter based on RF-PCT or RFML-C4.5, MLDF can achieve the best performance.
It indicates that MLDF has good flexibility.

\section{CONCLUSION} 
\label{conclusion}

In this paper, we first introduce the deep forest framework to multi-label learning and propose Multi-Label Deep Forest (MLDF). 
Because of the two measure-aware mechanisms, measure-aware feature reuse and measure-aware layer growth, our proposal can optimize different multi-label measures based on user demand, reduce the risk of overfitting, and achieve the best results on a bunch of benchmark datasets.

In the future, the efficiency of MLDF could be further improved by reusing some components during the process of forests training. Though our method can exploit high-order label correlations, we only show the second-order correlations in experiments. We will try to find a way to interpret how high-order correlations are used. Furthermore, we plan to embed extreme multi-label tree methods like FastXML \citep{prabhu2014fastxml} into MLDF and test the performance on extreme scale multi-label problems.

\bibliography{myRef}
\bibliographystyle{abbrvnat}

\newpage
\appendix
\end{document}